\documentclass{sig-alt-full}
\usepackage{algorithm}
\usepackage{algorithmic}
\begin{document}
\title{A Comparison of Evaluation Methods in Coevolution}

\numberofauthors{2}
\author{
\alignauthor 
Ting-Shuo Yo \\
       \affaddr{Utrecht University}\\
       \affaddr{PO Box 80.089}\\
       \affaddr{3508 TB Utrecht}\\
       \affaddr{The Netherlands}\\
       \email{tyo@cs.uu.nl}
\alignauthor 
Edwin D. de Jong \\
       \affaddr{Utrecht University}\\
       \affaddr{PO Box 80.089}\\
       \affaddr{3508 TB Utrecht}\\
       \affaddr{The Netherlands}\\
       \email{dejong@cs.uu.nl}
}
\maketitle

\begin{abstract}
In this research, we compare four different evaluation methods in coevolution on the Majority Function problem.  The size of the problem is selected such that an evaluation against all possible test cases is feasible.  Two measures are used for the comparisons, i.e., the objective fitness derived from evaluating solutions againt all test cases, and the objective fitness correlation (OFC), which is defined as the correlation coefficient between subjective and objective fitness.
The results of our experiments suggest that a combination of average score and weighted informativeness may provide a more accurate evaluation in coevolution.  In order to confirm this difference, a series of t-tests on the preference between each pair of the evaluation methods is performed.  The resulting significance is affirmative, and the tests for two quality measures show similar preference on four evaluation methods.
\end{abstract}


\section{Introduction}
Coevolution offers an approach to adaptively select tests for the evaluation of learners \cite{Hillis1990,rosin:1997:nmcc,juille:98,paredis:coevcomp,pagie:1998:ecct,dejong2004ideal}. Using coevolution, the evaluation function is adapted as part of the evolutionary process. This approach can be useful if the quality of individuals can be assessed using some form of {\em tests}. For such {\em test-based problems}, the identification of an informative set of tests can reduce the amount of required computation, while potentially providing more useful information than any static selection of tests. Since an adaptive test set can render evaluation unstable, an important question is how coevolution can be set up to be sufficiently reliable.

A recent insight in coevolution research is that the design of a coevolutionary setup should begin with a consideration of the desired {\em solution concept} \cite{ficici:phd}. A solution concept specifies which elements of the search space qualify as solutions and which do not. Examples of solution concepts include: Maximum Expected Utility (maximizing the expected outcome against a randomly selected opponent, which for uniform selection is equivalent to maximizing the average outcome against all opponents), the Pareto-optimal set resulting from viewing each test as a separate objective, and Nash-equilibria, in which no candidate solution or test can unilaterally deviate given the other candidate solutions and tests without decreasing its payoff.

For several of the main solution concepts used in coevolution, archive methods exist guaranteeing that when sufficiently diverse sets of new individuals are submitted to the archive, the archive will produce monotonically improving approximations of the solution concept. Recent examples of such archive methods are the Nash Memory \cite{ficici:phd,Ficici:2003:gecco}, which guarantees monotonicity for the Nash equilibrium solution concept; the IPCA algorithm, which guarantees monotonicity for the Pareto-optimal equivalence set \cite{dejong:coevj}; and the MaxSolve algorithm \cite{dejong:maxsolve}, which guarantees monotonicity for the Maximum Expected Utility solution concept.

While theoretical guarantees of monotonic progress are important, so far no bounds or guarantees regarding the improvement of the approximation to the solution concept over time are available. Thus, approximating the solution concept to a desired degree of accuracy may take an infeasible amount of time. An important current practical question is therefore: how can coevolutionary algorithms be set up such that their dynamics lead to quick improvement over time? By using such efficient algorithms as generators of new individuals and coupling them to monotic archives, thereby combining the guarantee of monotonic progress with efficiency, a principled approach to designing robust and efficient coevolution algorithms is obtained.

Efficiency in coevolutionary algorithms depends on selection, see e.g.~\cite{ficici:tec}, and evaluation. In this research we focus on evaluation. Our aim is to compare the efficiency, reflected in the improvement over time of an objective quality measure, that can be achieved using different coevolutionary evaluation methods. Since a main question is how sufficiently accurate evaluation may be achieved, the testing environment is chosen such that evaluating individuals on all tests is feasible; while this is not the case in practical applications of coevolution, this provides a possibility to compare evaluation methods with the maximally informative situation in which information about all possible tests is available. This setup permits investigating two important questions:
\begin{enumerate}
\item Given all information that may be relevant to evaluation, how can this information be used optimally?
\item Compared to evaluation based on all relevant information, how do different coevolutionary evaluation methods perform?
\end{enumerate}
In this paper, we focus on the second question. Four different coevolutionary evaluation methods are compared to each other and to the baseline of testing against all tests. The test problem is a small variant of the Majority Function test problem \cite{mitchell:1993} \cite{mitchell1994} \cite{juille:98} chosen such that evaluation against all test cases (initial conditions) is feasible. A new tool named the Objective Fitness Correlation (OFC) \cite{dejong:2007:ofc}, the correlation between the subjective and the objective fitness measures, is used to assess the evaluation accuracy of the different methods.  

The paper is structured as follows. In section 2 we discuss the evaluation methods and algorithms used in this research.  The design of experiments, parameters and performance measures are described in section 3.  The results are presented in section 4, and the discussions and concluding remarks are shown in section 5.


\section{Evaluation Methods}

In this section, we describe the evaluation methods conducted in this study.  Since we put our main focus on the test-based problems, we start with defining some terminologies and fitness measures.  Evaluation methods based on those fitness measures are introduced in the following sections. 

\subsection{Interaction matrix}
For a test problem, there are two sample spaces: one is the test cases, $T$, and the other is possible solutions, $S$.  The term {\it interaction} is defined as the result of letting one solution interact with one test case.  For the ease of analysis, an {\it interaction function}, $G(T,S)$, is designed to return one scalar outcome of the interaction between a pair of test case and a solution.  To simplify the following discussion, we assume that the interaction function returns a binary result which represents whether the solution succeeds in solving the test case or not.  An affirmative interaction is preferred by the solution, but is unfavorable to the test case.

By evaluating all solutions, ${\bf S}=\{S_j| j = 1,2,....,m\}$, against all test points, ${\bf T}=\{T_i| i = 1,2,....,n\}$, an interaction matrix, $I=\{I_{ij}|i = 1,....,n; j=1,....,m\}$, is obtained.  The sum of column $j$, $I_j$, of this interaction matrix represents the number of test cases that $S_j$ has solved, which is a common performance measure for a solution.  Similarly, the sum of row $i$, $I_i$, is the number of times this test case $T_i$ being solved, which represents the difficulty of the test case.

In the simplest case, only the interaction outcomes are considered, so that $-I_i$ can be used as the fitness of $T_i$ (i.e., the more difficult the better), and $I_j$ for $S_i$ (i.e., the more powerful the better).

\subsection{Distinctions}
Distinctions are defined as the ability to distinguish between good and bad solutions and may provide important information for selecting a proper set of test cases.  This concept was first proposed in \cite{ficici:2001:paretooptimality}, and here we follow the notation in \cite{dejong2004ideal} for our experiments.  Accordingly, a three-dimensional matrix $dist(T,S,S)$ is defined to represent "if test case $T_i$ distinguishes $S_k$ from $S_l$", or mathematically:
\begin{equation}
dist(T_i,S_k,S_l) \Leftrightarrow G(T_i,S_k) > G(T_i,S_l)
\label{def:dist}
\end{equation}
where $T_i \in {\bf T}$ and $S_k, S_l \in {\bf S}$.\\
The number of distinctions that test case $T_i$ has made may be represented by summing up all $m^2$ solution pairs in $dist$, i.e., $dist_T(T_i) = \sum_{kl}dist(i,k,l)$.  This value represents the ability of a test case to maintain diversity of the solutions, and it may also be considered as a fitness measure for test cases.

Although we explain the concept of distinction in terms of ``selecting a proper set of test cases'', the same procedure can also be applied to solutions.  This yields the distinction of solutions, and is also a reasonable fitness measure for evaluating solutions.  With those measures, four different evaluation methods are defined as follows.

\subsection{Four evaluation methods}
For each evaluation method, we define the fitness for test cases $F(T)$ and solutions $F(S)$ as follows.
\subsubsection{Average score}
For a solution, the average-score is defined as ``the proportion of test points it has succeeded in solving.''  And for a test point, this value is represented by "the proportion of solutions that it has failed".  With an interaction matrix $I(i,j)$ as described earlier, we define:
\begin{eqnarray}
F_{AS}(T_i) &=& 1 -\sum_{j=1}^{m} I(i,j) / m, \nonumber \\
F_{AS}(S_j) &=& \sum_{i=1}^{n} I(i,j) / n
\label{eqn:ave-score}
\end{eqnarray}
The application results of $F_{AS}$ on test cases and solutions are both between $[0,1]$, and a higher value represents a better performance.

\subsubsection{Weighted score}
Instead of giving each interaction an equal weight, the average scores described above may be used as weights for each test case and solution.  That is to say, a test case earns more credits by failing a more powerful solution, and a solution gets a higher score when it succeeds in a tough test case.  The mathematical expression of this evaluation can be represented as follows:
\begin{eqnarray}
W_T(T_i) &=& \left(\sum_{j=1}^{m} I(i,j)\right)^{-1},\nonumber \\
W_S(S_j) &=& \left(\sum_{i=1}^{n} I(i,j)\right)^{-1},\nonumber \\
W'_T(T_i) &=& \frac{W_T(T_i)}{\sum_{i=1}^{n} W_T(T_i)}, \nonumber \\
W'_S(S_j) &=& \frac{W_S(S_j)}{\sum_{j=1}^{m} W_S(S_j)}, \nonumber \\
F_{WS}(T_i) &=& 1 - \sum_{j=1}^{m} W'_S(S_j) \cdot I(i,j),\nonumber \\
F_{WS}(S_j) &=& \sum_{i=1}^{n} W'_T(T_i) \cdot I(i,j)
\label{eqn:wei-score}
\end{eqnarray}
Weights are checked to avoid being divided by zero and normalized to ensure the weighted scores are ranged from $0$ to $1$.  Hence $F_{WS}$ also returns a value between $[0,1]$, and the higher values are more preferable.  This definition is similar in spirit to the niching methods of Rosin \cite{rosin:1997:nmcc} and Juille \cite{juille:1996b}, though their methods are not considered while we developed the weight function.

\subsubsection{Average informativeness}
The informativeness of a test measures the amount of information it provides about a given set of candidate solutions.
In \cite{bucci:foga}, a definition for informativeness based on the incomparable and equal elements of the order induced by a test is provided. Here, we measure the informativeness of a test based on Ficici's notion of distinctions \cite{ficici:2001:paretooptimality}. Since each distinction a test makes contributes to its informativeness, and since the set of all possible distinctions is sufficient to provide ideal evaluation \cite{dejong2004ideal}, we measure the informativeness of a test as the normalized number of distinctions it makes.
In this study, we define the {\it distinction score}, $D$, as ``the number of distinctions one test case (solution) makes.''  This value can be derived from the distinction matrix described in (\ref{def:dist}), and represents the informativeness.  For the convenience of further computation, we normalize the distinction score with its maximum and minimum values, which are returned by the function $Max(D)$ and $Min(D)$.  Because the informativeness represents the individual's ability to maintain diversity of the other population, we want to integrate it with the average scores to form the fitness. Here we simply use a linear combination of two scores, with weights of $(0.3,0.7)$ for distinction and average score, respectively.  These weights are based on the observation that the average scores are generally lower than the distinction scores due to the computational scheme we used, as well as experiences from the pilot experiments. The average informativeness is therefore defined as the linear combination of the normalized distinction score and the average score.

\begin{eqnarray}
D_T(T_i) &=& \sum\sum_{k,l \in [1,m]}dist(T_i,S_k,S_l), \nonumber \\
D_S(S_j) &=& \sum\sum_{p,q \in [1,n]}dist(S_j,T_p,T_q), \nonumber \\
D_T'(T_i) &=& \frac {D_T(T_i)-Min(D_T(T_i))}
                      {Max(D_T(T_i))-Min(D_T(T_i))}, \nonumber \\
D_S'(S_j) &=& \frac {D_S(S_j)-Min(D_S(S_j))}
                      {Max(D_S(S_j))-Min(D_S(S_j))}, \nonumber \\
F_{AI}(T_i) &=& 0.3\cdot D_T'(T_i) + 0.7\cdot F_{AS}(T_i),\nonumber \\
F_{AI}(S_j) &=& 0.3\cdot D_S'(S_j) + 0.7\cdot F_{AS}(S_j)
\label{eqn:ave-info}
\end{eqnarray}

\subsubsection{Weighted informativeness}
Similar to the weighted score, each distinction that has been made can be weighted differently when the distinction score is derived.  For example, the distinction made by all test cases, $\sum_{i=1}^{n} dist(T_i,S_k,S_l)$, provides the information of ``how many test cases have made the distinction on $S_k$ and $S_l$, such that $G(T_i,S_k) > G(T_i,S_l)$.''  The inverse of this value can be used as the weight of this particular distinction, so that when more test cases can make this distinction, the less worthy this distinction is.  This operation can also be applied to solutions, and the weighted informativeness can be defined mathematically as:

\begin{eqnarray}
&W_d(S_k,S_l) &= \left(\sum_{i=1}^{n} dist(T_i,S_k,S_l)\right)^{-1},\nonumber \\
&W_d(T_p,T_q) &= \left(\sum_{j=1}^{m} dist(S_j,T_p,T_q)\right)^{-1},\nonumber \\
&D_T(T_i) &= \sum\sum_{k,l \in [1,m]} W_d(S_k,S_l) 
               \cdot dist(T_i,S_k,S_l), \nonumber \\
&D_S(S_j) &= \sum\sum_{p,q \in [1,n]} W_d(T_p,T_q) 
                \cdot dist(S_j,T_p,T_q), \nonumber \\
&D_T'(T_i) &= \frac {D_T(T_i)-Min(D_T(T_i))}
                      {Max(D_T(T_i))-Min(D_T(T_i))}, \nonumber \\
&D_S'(S_j) &= \frac {D_S(S_j)-Min(D_S(S_j))}
                      {Max(D_S(S_j))-Min(D_S(S_j))}, \nonumber \\
&F_{WI}(T_i) &= 0.3\cdot D_T'(T_i) + 0.7\cdot F_{AS}(T_i),\nonumber \\
&F_{WI}(S_j) &= 0.3\cdot D_S'(S_j) + 0.7\cdot F_{AS}(S_j)
\label{eqn:wei-info}
\end{eqnarray}

In the following discussion, the four evaluation methods described above are referred as {\bf AS}, {\bf WS}, {\bf AI,} and {\bf WI}, respectively.

\subsection{Algorithms for experiments}
There are two main algorithms used in this study, i.e., a single population genetic algorithm (GA) and coevolution (CO).  The former uses only one single population for solutions and evaluates the population on all possible test cases to obtain the interaction defined earlier.  Afterward, the average scores given by $F_{AS}$ are used as the fitness in the GA.  The results of this algorithm are used as the baseline for comparisons.  The second algorithm uses coevolution between test cases and solutions, and all four evaluation methods are tested.

Algorithm \ref{alg:1pop-ga} and \ref{alg:coevolution} describe GA and CO, respectively.  Functions used in the algorithms, e.g., INTERACTION, EVALUATE, SELECT and BREED, are specific to problems and experiments.  The general concepts of INTERACTION and four implementations of EVALUATE are already discussed in this section.  The SELECT and BREED together define the selection, reproduction and replacement cycle in a generation.  The choice of these two functions is problem specific and they are specified in the experimental design section.

\begin{algorithm}
\caption{\hspace{8pt} Single Population GA}
\label{alg:1pop-ga}
\begin{algorithmic}
  \STATE $TC \leftarrow$ all $n$ test cases
  \STATE $Sol(0) \leftarrow$ randomly generate $m$ solutions
  \WHILE {$t <$ MAX\_GENERATION}
    \STATE $I \leftarrow$ INTERACTION($TC,Sol(t)$)
    \STATE $Sol(t) \leftarrow$ EVALUATE $Sol(t)$ with $F_{AS}$
    \STATE $Sol'(t)  \leftarrow$ SELECT $Sol(t)$
    \STATE $Sol(t+1) \leftarrow$ BREED $Sol'(t)$
    \STATE $t = t+1$ 
  \ENDWHILE
  \RETURN $Sol(t)$
\end{algorithmic}
\end{algorithm}

\begin{algorithm}
\caption{\hspace{8pt} Coevolution}
\label{alg:coevolution}
\begin{algorithmic}
  \STATE $TC(0)  \leftarrow$ randomly generate $n$ test cases
  \STATE $Sol(0) \leftarrow$ randomly generate $m$ solutions
  \WHILE {$t <$ MAX\_GENERATION}
    \STATE $I \leftarrow$ INTERACTION($TC(t),Sol(t)$)
    \STATE $TC(t)  \leftarrow$ EVALUATE $TP(t)$
    \STATE $Sol(t) \leftarrow$ EVALUATE $Sol(t)$
    \STATE $TC'(t)   \leftarrow$ SELECT $TC(t)$
    \STATE $Sol'(t)  \leftarrow$ SELECT $Sol(t)$
    \STATE $TC(t+1)  \leftarrow$ BREED $TC'(t)$
    \STATE $Sol(t+1) \leftarrow$ BREED $Sol'(t)$
    \STATE $t = t+1$ 
  \ENDWHILE
  \RETURN ($TP(t), Sol(t)$)
\end{algorithmic}
\end{algorithm}

\section{Experimental Setup}
In this section, the settings of our experiments are described.  The software used in this research is implemented as an extension of ECJ \cite{ECJ15:website}, which is developed by George Mason University's Evolutionary Computation Laboratory.  All simulations use the basic evolutionary loop provided by ECJ, plus the evaluation methods and problem-specific functions in the extension.  The design of the experiments, the parameters used in the test problem, and the performance measures are introduced as follows. 

\subsection{Design of experiments}
As mentioned in the previous section, two algorithms and four evaluation methods are used in this study. Table \ref{tab:exp-setting} shows the design of our experiments.  Experiment 1 is a combination of the single population genetic algorithm and the average score evaluation method.  This GAAS evaluates the solutions against all possible test cases and serves as the baseline experiment.  Experiment 2, 3, 4, and 5 are four different evaluation methods combined with the coevolution algorithm.  Each experiment is run for ten times with ten different random seeds.

\begin{table}
\begin{center}
\begin{tabular}{lccccc}
\hline
Exp No            & 1  & 2  & 3  & 4  & 5 \\ 
\hline
Algorithm         & GA & CO & CO & CO & CO \\ 
Evaluation Method & AS & AS & WS & AI & WI \\ 
Number of Runs    & 10 & 10 & 10 & 10 & 10 \\ 
\hline
\end{tabular}
\caption{Design of experiments.}
\label{tab:exp-setting}
\end{center}
\end{table}

\subsection{Parameters for the test problem}
In this study, we perform the experiments on the majority function problem.  This problem is also known as the one dimensional cellular automata problem.  Mitchell and his colleagues have detailed discussion on this problem in \cite{mitchell:1993} and \cite{mitchell1994}.  Two major parameters are used for this problem, i.e., the radius of the neighbourhood ($r$) and the size of the one dimensional lattice ($N$).  Boolean vectors are used to represent both the initial conditions (test cases) and the rules (solutions).  In order to evaluate against all test points, we chose $r=2$ and $N=9$.  Other parameters, e.g., the size of populations, elitism, crossover type, and mutation probability are selected based on Mitchell's work \cite{mitchell1994}, and are summarized in Table \ref{tab:exp-par-maj}.

In the coevolution experiments (2, 3, 4, and 5), a symmetric setup is used, i.e., the same set of EVALUATE, SELECT and BREED functions are used for both population of test cases and the population of solutions.  However, there is one exception: the population size.  Because there are only 512 possible test cases in total, the population size for test cases is set as 64 in the convenience of comparison, while the population size for solutions is set as 100.  Accordingly, the number of elites for two populations are also different, while their proportions to the size of populations are the same (20\%).  The initial populations are both randomly created and are both checked to ensure there are no redundent individuals in the begining of the experiments.
\begin{table}
\begin{center}
\begin{tabular}{ll}
\hline
Parameter & Value \\ 
\hline
MAX\_GENERATION & 200 \\
Size of population & 64/100 \\
Elitism & 12/20 (20\%) \\
Selection & linear rank selection \\ 
Crossover & one point crossover \\
Mutation rate & 0.01 \\ 
\hline
\end{tabular}
\caption{Parameters used for the majority function problem.}
\label{tab:exp-par-maj}
\end{center}
\end{table}
  


\subsection{Performance measures}
In order to compare all experiments fairly, an objective performance measure is required.  Hence we define the {\it subjective fitness} and {\it objective fitness} separately.  The {\it subjective fitness} is the fitness value returned by the EVALUATE function used in the coevolution algorithm.  In this study, the {\it objective fitness} is defined as the fitness used in experiment 1, i.e., to evaluate the average scores against all possible test cases.  For each generation in experiment $2\sim5$, both the subjective and the objective fitness are recorded.  All experiments are compared with the their objective fitness, based on the number of interactions.  In coevolution experiments, the number of test cases is $1/8$ of that in the GAAS experiment, therefore during the analysis we compare the objective fitness of the coevolution experiments every $8$ generations to the fitness of the controlled experiments every generation.  Since these values are all derived from exhaustive testing, the progress over interactions may be seen as the performance of each method.

In addition to the progress of the fitness, the correlation coefficients between subjective and objective fitness are also computed.  This correlation is defined as the {\bf Objective Fitness Correlation}, OFC, in \cite{dejong:2007:ofc} as a new objective measure for evaluating coevolutionary algorithms.  In our experiments, this measure is collected for each generation of every run.  Since OFC is always $1$ in the baseline experiment, we only compare OFC among the four evaluation methods with coevolution.  Comparisons are made with these two measures averaged upon ten independent runs.

\section{Results}

\subsection{Best objective fitness}
\begin{figure} 
 \centering 
 \includegraphics[scale=0.45]{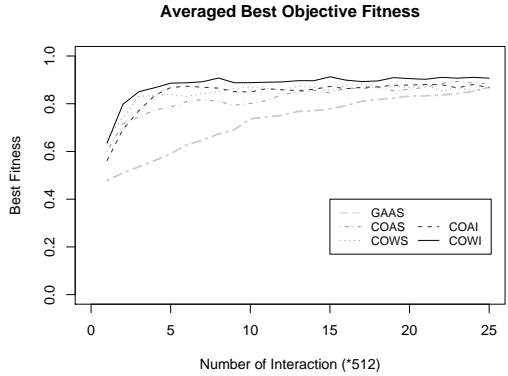} 
 \caption{The best objective fitness of solutions as a function of the number of interactions}
 \label{fig:mj-objf} 
\end{figure}

Figure \ref{fig:mj-objf} shows the best objective fitness of the solutions averaged over 10 runs.  As illustrated in the figure, the baseline experiment is outperformed by all coevolution experiments, especially when the number of total interactions is small.  Among four evaluation methods with coevolution, the weighted informativeness shows a constantly better performance than others.

A series of paired, one-tailed t-tests are performed to examine the significance of the differences between each evaluation methods.  The best fitness of each generation is averaged over 10 runs, and is paired with the same quantity of another evaluation method for the t-test. The results of t-tests shows significance on WI $>$ AI, WI $>$ WS, WI $>$ AS, AI $>$ AS, and WS $>$ AS (with p-values $ < 1 \times 10^{-10}$), but not on AI $\ne$ WS with $\alpha = 0.05$.
These results are summarized in table \ref{tab:t-test-objf}.

\subsection{Objective fitness correlation}
\begin{figure} 
 \centering 
 \includegraphics[scale=0.45]{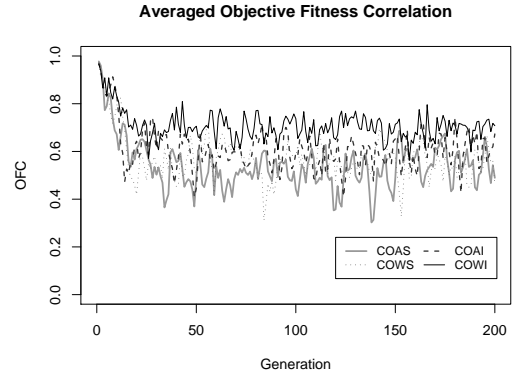} 
 \caption{The OFC for experiment COAS, COWS, COAI and COWI.}
 \label{fig:mj-ofc} 
\end{figure}

The OFCs for different evaluation methods with coevolution are shown in figure \ref{fig:mj-ofc}.  The values shown in the figure are averages of 10 independent runs, and therefore the fluctuations of OFCs are already smoothed.  In a single run of one experiment, the OFC can be negative for some generations.  

In all experiments, the OFC always starts from a high value.  This is because the initial populations are randomly generated, and this can be seen as they are sampled from the set of all possible cases with a uniform distribution.  Ideally, this random sampling may create a set well represents the original search space, i.e., the all possible cases, and results in a higher correlation between the subjective and the objective fitness.  As the coevolution proceeds, the populations move toward certain direction rather than a uniformly random sampling, and hence the OFC decreases over generations.  However, as argued in \cite{dejong:2007:ofc}, the OFC may still remain as a performance measure for comparing different evaluation methods in a coevolutionary algorithm.

As demonstrated in figure \ref{fig:mj-ofc}, the weighted informativeness shows a constantly higher OFC through generations than other methods, while other evaluation methods do not show clear differences between one another due to the fluctuations.

A set of t-tests is employed to verify the difference in OFC between each pair of evaluation methods.  The results show significance on all of following relations: WI $>$ AI, WI $>$ WS, WI $>$ AS, AI $>$ WS, AI $>$ AS, and WS $>$ AS.  The preferences for WI against other methods are very significant, with p-values $ < 1 \times 10^{-15}$, which is the limit of precision in the computing software.
These results are summarized in table \ref{tab:t-test-ofc}.

\begin{table}
\begin{center}
\begin{tabular}{cccc}
\hline
P-value for test &&&\\
on i $<$ j  & WS & AI & WI \\ 
\hline
AS & $1.4\times 10^{-21}$ & $9.9\times 10^{-15}$ & $1.9\times 10^{-68}$ \\
WS &                      & $9.8\times 10^{-1}$ & $7.2\times 10^{-81}$ \\
AI &                      &                      & $1.5\times 10^{-64}$ \\
\hline
\end{tabular}
\caption{Significant level in paired t-tests for objective fitness among experiment COAS, COWS, COAI and COWI.   A p-value smaller than 0.05 is usually considered as significant.}
\label{tab:t-test-objf}
\end{center}
\end{table}

\begin{table}
\begin{center}
\begin{tabular}{cccc}
\hline
P-value for test &&&\\
on i $<$ j  & WS & AI & WI \\ 
\hline
AS & $4.2\times 10^{-5}$ & $6.5\times 10^{-21}$ & $2.9\times 10^{-69}$ \\
WS &                     & $4.6\times 10^{-9}$  & $3.4\times 10^{-59}$ \\
AI &                     &                      & $1.1\times 10^{-39}$ \\
\hline
\end{tabular}
\caption{Significant level in paired t-tests for OFC among experiment COAS, COWS, COAI and COWI.}
\label{tab:t-test-ofc}
\end{center}
\end{table}

\section{Discussion}

\subsection{Evaluating with the weighted informativeness}
\begin{figure} 
 \centering 
 \includegraphics[scale=0.45]{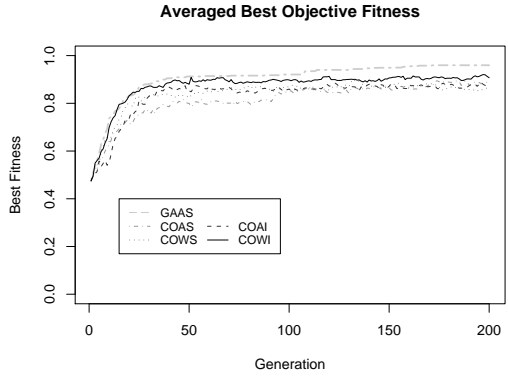} 
 \caption{The objective fitness over generations for experiment GAAS, COAS, COWS, COAI and COWI.}
 \label{fig:mj-objf-gen} 
\end{figure}

The results of our experiments suggest that in coevolution, a combination of performance (average score) and diversity (weighted distinction) can achieve the accuracy of full evaluation with less computational cost.  This statement holds for both conditions when considering the improvement of an objective quality measure over time or the correlation between the subjective fitness and objective measure.  This advantage is even clearer if we compare the objective fitness according to the generation instead of the number of interactions.  As shown in figure \ref{fig:mj-objf-gen}, the WI evaluation method with coevolution progresses as fast as evaluating against all possible test cases, while other evaluation methods are apparently slower than the baseline.

In our study, the OFC is used as a quality measure in addition to the objective fitness.  If the preference orderings of evaluation methods are considered, this measure provides information similar to the objective fitness.  However, the difference between WS and AI shows significance in OFC but not in the objective fitness.  This implies that two measures may contain different information, and a detailed discussion on OFC can be found in \cite{dejong:2007:ofc}.

\begin{figure} 
 \centering 
 \includegraphics[scale=0.45]{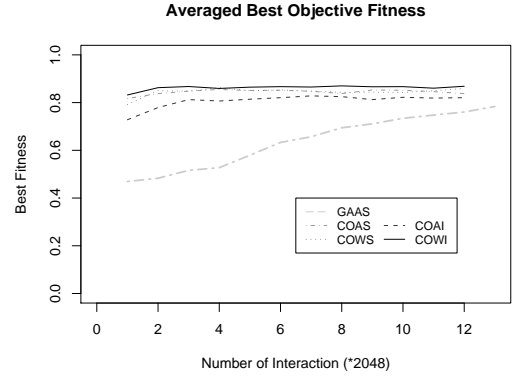} 
 \caption{The averaged best objective fitness for the $N=11$ majority function problem.}
 \label{fig:mj-r2n11-objf} 
\end{figure}

\begin{figure} 
 \centering 
 \includegraphics[scale=0.45]{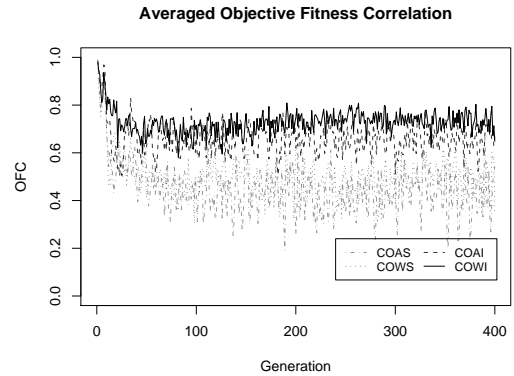} 
 \caption{The averaged OFC for the $N=11$ majority function problem.  The upper lines are AI and WI, and the lower lines are AS and WS.}
 \label{fig:mj-r2n11-ofc} 
\end{figure}

\subsection{The majority function problem with different sizes}
This research is the first time the OFC is calculated on a real problem, but in exchange we are only able to experiment on problems with a small number of possible test cases in total.  The majority function problem with different sizes ($N=7$ and $N=11$) are also tested.  The results are similar to the finding presented in the previous section, and the experiments of $N=11$ are summarized in figure \ref{fig:mj-r2n11-objf} and \ref{fig:mj-r2n11-ofc}.  From these figures it is shown that OFC can distinguish AI and WI from AS and WS while the objective fitness can not.  This argument is consistent with the results of t-tests (not shown): t-tests on the objective fitness show no significant difference between pairs among AS, WS, and AI, while those on OFC show the same preference as in the $N=9$ experiments.

\subsection{The parity problem}
In addition to the majority function problem, the same comparison is also performed on the $n$-odd-parity problems.  Although Lee, Xu, and Chau \cite{lee:2001:ppca} have proposed that the boolean parity problem may be transformed into a cellular automata problem, here we use Koza's Genetic Programming approach in \cite{koza:1992:gp}.

\begin{figure} 
 \centering 
 \includegraphics[scale=0.45]{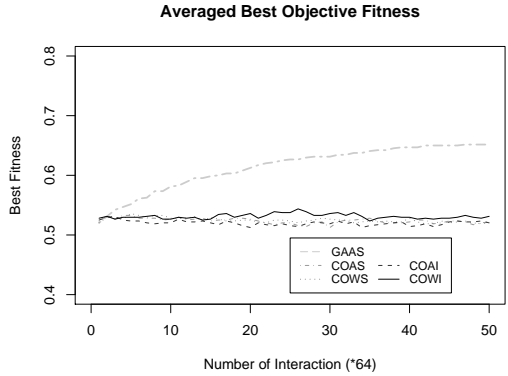} 
 \caption{The averaged objective fitness for the $6$-odd-parity problem.}
 \label{fig:par-n06-objf} 
\end{figure}

\begin{figure} 
 \centering 
 \includegraphics[scale=0.45]{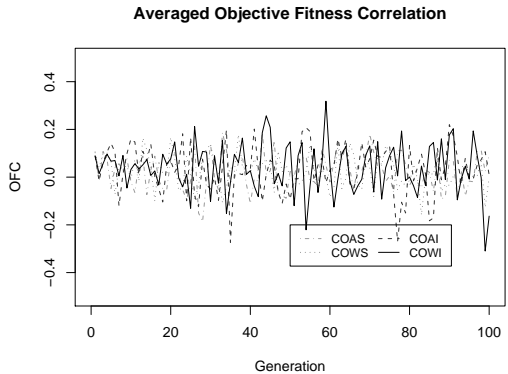} 
 \caption{The averaged OFC for the $6$-odd-parity problem.}
 \label{fig:par-n06-ofc} 
\end{figure}

Figure \ref{fig:par-n06-objf} and \ref{fig:par-n06-ofc} show the results for the $6$-odd-parity problem.  As shown in figure \ref{fig:par-n06-objf}, the WI still performs the best among four evaluation methods in coevolution, and  this is also confirmed by the t-tests (not shown).  However, we also find some disagreement between results from the majority function problem and the parity problem.  First, in the parity problem, the single population GA outperforms all CO experiments in this genetic programming approach, while figure \ref{fig:mj-objf} shows the opposite in the majority function problem.  Second, OFC in the parity problem shows no significant pattern (as shown in figure \ref{fig:par-n06-ofc}), while the OFC distinguishes different evaluation methods better than the objective fitness in the majority function problem.  Finally, the t-tests hardly show any significant difference among AS, WS, and AI in the parity problem.

A detailed analysis on the output has been done and suggests a few possible reasons for the disagreement.  First, solving the parity problem with the genetic programming approach is not a symmetric test-based problem in nature.  That is to say, while the test cases are represented with boolean vectors as they are in the majority function problem, the solutions for the parity problem are represented with tree-like structures.  As a result of this asymmetry, the same evolutionary operator (e.g., selection methods, mutation rate, crossover methods, etc.) may have different effects on two populations, and hence a symmetric setting for coevolution is not suitable.  Second, the initial population of the tree-like solutions is not randomly selected from ``all possible solutions''.  In Koza's approach, the sizes of the solutions start from smaller values and grow over generations.  This ``biased sampling'' may explain the disfavor of OFC in the parity problem.
Finally, since Koza's approach has been studies for several years, the evolutionary operators and parameters are already well tuned.  We believe that the disagreement may be reduced by developing proper asymmetric coevolution schemes.

Despite the disagreement, the results of the parity problem still shows a favor to WI in coevolution.

\section{Conclusion}
In this research, we compare four different evaluation methods in coevolution on a test-based problem.  Two measures are used for the comparisons among average score (AS), weighted score (WS), average informativeness (AI), and weighted informativeness (WI).  In addition to an objective quality measure, the objective fitness correlation (OFC) is also computed.

The experimental results show a strong preference on WI, which suggest that a combination of the performance and the ability to create distinctions may provide more accurate evaluation in coevolution.  The resulting significance from t-tests show a similar preference when two quality measures are used, separately.  This study also uses the recently proposed OFC to evaluate the accuracy of coevolutionary evaluation methods on a concrete test problem.  

Although we have shown the advantages of using WI in coevolution, the way we combine these two measures is simply using a weighted summation.  It may worth exploring more sophisticated methods to fuse this information together.  Currently a multi-objective approach is in progress, and this may lead to a more detailed investigation on how to use both measures in coevolutionary algorithms.



\bibliographystyle{abbrv}   
\bibliography{coevo}        
\balancecolumns

\end{document}